\documentclass[review]{elsarticle}
\usepackage{graphicx}
\usepackage{amsmath}
\usepackage{hyperref}
\usepackage{subfigure}
\usepackage{amsfonts}
\usepackage{multirow}
\usepackage[ruled,vlined]{algorithm2e}
\usepackage[utf8]{inputenc}
\usepackage[T1]{fontenc}

\begin{document}
\begin{frontmatter}

\title{Believe The HiPe: Hierarchical Perturbation for Fast, Robust, and Model-Agnostic Saliency Mapping}

\author[1]{Jessica Cooper\fnref{a}}
\author[1]{Ognjen Arandjelović}
\author[1]{David J Harrison}
\address[1]{University of St Andrews}
\fntext[a]{jmc31@st-andrews.ac.uk}


\begin{abstract}

Understanding the predictions made by Artificial Intelligence (AI) systems is becoming more and more important as deep learning models are used for increasingly complex and high-stakes tasks. Saliency mapping -- a popular visual attribution method -- is one important tool for this, but existing formulations are limited by either computational cost or architectural constraints. We therefore propose Hierarchical Perturbation, a very fast and completely model-agnostic method for interpreting model predictions with robust saliency maps. Using standard benchmarks and datasets, we show that our saliency maps are of competitive or superior quality to those generated by existing model-agnostic methods -- and are over 20$\times$ faster to compute.

\end{abstract}
\end{frontmatter}

\section{Introduction}

As deep learning is applied to increasingly high-stakes domains, the ability to accurately explain model predictions in a human-interpretable way is becoming more and more important~\cite{Adadi2018-qm,bai2021explainable}. Developing tools to understand whether, for example, a model classifies a biopsy image as malignant due to cell morphology, or due to a smudge on the slide, is crucial if we are to see safe, widespread adoption of powerful AI techniques~\cite{jiang2021learning,valsson2022nuances,barata2021explainable,hryniewska2021checklist}.

It is important to note that interpretability research, also known as XAI (eXplainable Artificial Intelligence) research, is relatively nascent and as such is as yet somewhat ill-defined: the terms ‘interpretability’ and ‘explainability’ are typically used interchangeably, and there is a lack of consensus regarding formal definitions of either or both of them~\cite{Murdoch2019-bq, Linardatos2020-zd, Miller2019-zm}. This problem is beyond the scope of our work here, so for our purposes we simply define them as does Miller: ``the degree to which a human can understand the cause of a decision''~\cite{Miller2019-zm}. 

Which XAI methods are best suited to a given model is primarily determined by the architecture of that model. Some simpler models (sometimes called `white-box' models) are considered intrinsically explainable, such as small decision trees and regression models~\cite{Rudin2019-dj} -- however, this restriction of complexity severely limits their application. For example, we might use gradient descent to fit a linear regression model to predict patient mortality $m$ based on clinical data $x_{0...n}$ -- such as age $x_{0}$, immune cell count $x_{1}$, and cancer stage $x_{2}$ etc: $m = x_{0}\theta_{0} + x_{1}\theta_{1} + x_{2}\theta_{2} + ...\theta_{n}$. This hypothetical model would be intrinsically interpretable -- we already understand the input features (i.e.\ we know what 'immune cell count' is, and what it means), and we can simply inspect the learned coefficient vector $\theta$ to see how the model combines them to make a prediction.

This inherent interpretability cannot be relied upon if input features interact to produce the output prediction in complex, non-linear ways, and so require complex, non-linear models, as is often the case in modern applied machine learning -- and so we turn to more complex models which are able to capture these complex relationships, but which are \textit{not} inherently interpretable. Taking the hypothetical mortality predictor above as an example, we might find that linear regression is unable to adequately capture the relationship between input features and mortality. Instead, we might use a neural network for its greater expressive power. However, even small neural networks cannot be interpreted in the same straightforward way - instead of $m = \sum_{i=0}^{n}x_{i}\theta_{i}$, where the learned parameters $\theta$ consist of a single vector with each element corresponding to the learned weight of an input feature, we are faced with $m = a(\theta^{(2)} + \beta^{(2)}a(\theta^{(1)} + \beta^{(1)}a(x\theta^{(0)} + \beta^{(0)})))$, where \textit{a} is some non-linearity, $\theta$ are the learned weights, $\beta$ is the bias at each layer (denoted$^{0...n-1}$ for a network with \textit{n} layers -- in this case, three). XAI methods therefore aim to explain or interpret the predictions of these kinds of more complex models -- that is, models that we cannot understand by merely inspecting their parameters (often termed `black-box' models~\cite{johansson2022rule,muddamsetty2022visual}).

Interpretability methods fall into two broad groups: global (what has the model learned from the dataset?); and local (what has caused the model to produce this particular output for an individual sample?). Global methods aim to explain how a model makes predictions holistically, identifying which features are important across the whole dataset and how model outputs are distributed based upon the data, learned parameters and model architecture. For complex models this is hard to achieve in a way that is interpretable by humans (for example, it is very difficult to image a hyperplane with more than three dimensions), so in practice most global methods focus only on some parts of the model -- such as the learned weights in a convolutional layer, to explain which features the model has learned to identify. Feature visualisation~\cite{Molnar2021-ce} is one example of this kind of global interpretability method, in which a single input sample is optimised to maximise the output of a particular node, layer, or logit. This optimised input can then be inspected in order to identify what kinds of features in the input that particular node, layer, or logit has learned to respond to. 

Local methods, by contrast, are concerned only with individual data samples and aim to explain why the model produces the output that it does \textit{given} that particular input~\cite{Molnar2021-yy}. These methods are typically much easier for humans to parse as they answer questions which can be easily visually interpreted -- ``Which elements of this sample influenced the output most?''.

In this work we are concerned with this kind of local explanation -- also known as \emph{attribution} -- that is, the identification of the portions of a given input that are most important in determining a model's output. One way of doing this in an easily interpretable way is by generating a saliency map -- a heat-map assigning colour or brightness to regions of the input according to how much each region contributed to the output. Saliency visualisation of this kind is widely used in machine learning, particularly in image classification tasks that often require the use of large and complex neural networks which are troublesome to interpret otherwise. Saliency maps are intuitive to interpret, and are typically used to validate that models are learning to identify sensible features to make predictions and identify biases, which is not only crucial for safety as AI is increasingly applied to high stakes domains such as medicine and autonomous vehicles, but also a valuable tool in increasing trust in and adoption of these powerful technologies.

Existing saliency mapping methods fall into two broad categories -- those which are \emph{model-specific}~\cite{kook2022deep}, and depend on access to the structure and internal state of the trained model, and those which are \emph{model-agnostic}~\cite{rio2020understanding}, and only require access to the input and output of the model. Model-specific methods are typically much more efficient, as they work by inspecting the internal state of the model and as such, do not require iteration over different permutations of the input, but can only be used when the trained architecture of the model in question is both accessible, and of a specific type (typically a deep convolutional neural network), which limits their application~\cite{Molnar2021-yy}. They cannot be used at all for models in which the internal state and structure is inaccessible.

In contrast, model-agnostic methods work by perturbing the input and inspecting the change in model output to determine the perturbed region's importance, irrespective of the internal state of the model -- and so can be used for any type of model at all, including ensemble methods which could combine both white- and black-box models, and so have wide-ranging applications. However, existing techniques of this kind are slow, as they build up an empirical estimation of region importance through many iterations -- and moreover, require a number of hyperparameters to be specified -- the optimal values of which are difficult to know ahead of time for new datasets, necessitating computationally expensive heuristic tuning.

There are a number of important AI applications where explicability is very important, but where the individual data samples are too large for existing model-agnostic saliency methods to be computationally feasible. Additionally, in many cases the network architecture is inaccessible or unsuitable for the application of faster gradient- and activation-based model-specific explanatory methods, or those methods would result in saliency maps too coarse for the task at hand. These might be things like healthcare triage using ensemble architectures taking both clinical data and high resolution medical imaging; very high resolution image inputs in general; legal tasks using very large textual or tabular datasets, or autonomous vehicle decision-making combining video, sensor and time-series input, to name a few~\cite{Manikandan2020-yw, Molnar2021-yy}.

We therefore contribute a novel model-agnostic saliency mapping method, Hierarchical Perturbation (HiPe) which is fast, robust, and completely model-agnostic, and evaluate it on the commonly-used MSCOCO and VOC2007 validation and test datasets, using the pointing game benchmark~\cite{Zhang2016-vm} and the causal insertion/deletion metrics~\cite{Petsiuk2018rise}. We show that HiPe is over 20$\times$ faster than existing model-agnostic methods, while achieving comparable or superior performance on these benchmarks.

\section{Related Work}

\begin{figure}
    \centering
    \includegraphics[width=1\textwidth]{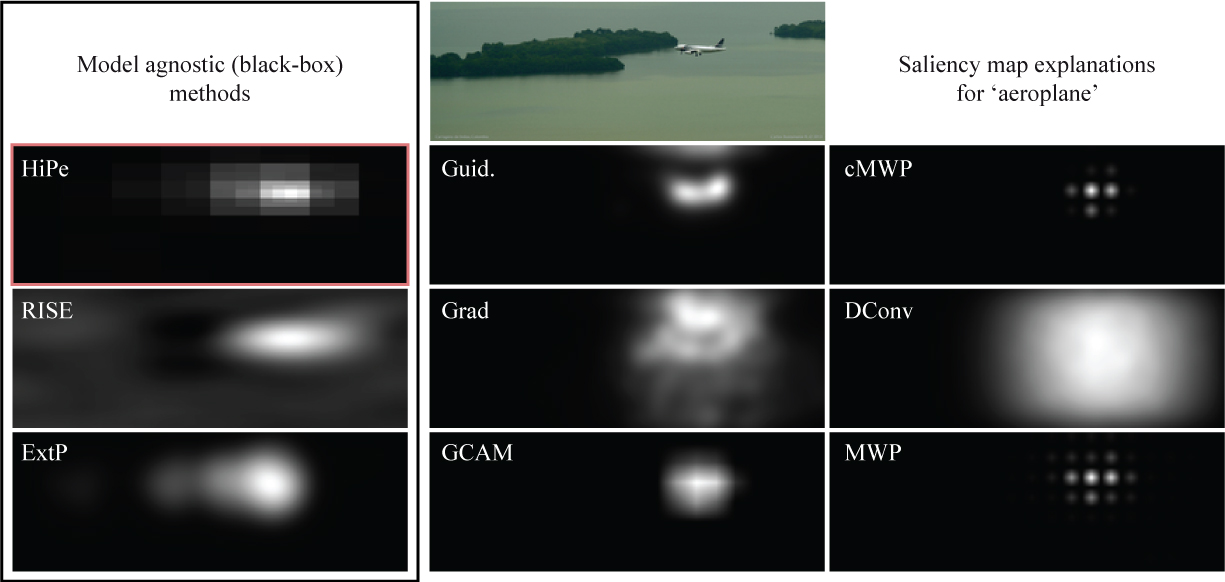}
    \caption{Examples of saliency maps generated by: Hierarchical Perturbation (HiPe, our proposed method); RISE~\cite{Petsiuk2018rise}; Extremal Perturbation (ExtP)~\cite{Fong2019-wy}; Guided Backpropagation (Guid.)~\cite{Springenberg2014-rv}; Gradient (Grad)~\cite{Simonyan14a}; Grad-CAM (GCAM)~\cite{Selvaraju2016-vq}; Contrastive Excitation Backpropagation (cMWP)~\cite{Zhang2016-vm}; Deconvnet (DConv)~\cite{Zeiler2013-lm}; and Excitation Backpropagation (MWP)~\cite{Zhang2016-vm}.}
    \label{fig:sal_map_examples}
\end{figure}

\subsection{Model-Specific Saliency Mapping Methods}

Model-Specific methods typically leverage the convolutional network architecture to visualise explanations by using gradients, activations, or some combination of the two~\cite{Selvaraju2016-vq, Simonyan14a, Zhang2016-vm, Mahendran2016-hf, Zhang2016-vm}. They are efficient, but the resolution of the maps they generate is architecture dependent, and some (guided backprop and deconvnet, in particular) have proven unreliable -- in some cases no better than edge detection~\cite{Adebayo2018-mj, Schneider2020-jx, Nie2018-zw}. As pointed out by Fong et al.~\cite{Fong2019-vk}, they are also fundamentally ungrounded in \emph{what} makes some region of the input more or less salient -- their explanatory power is assessed a posteriori. At present, even the most successful methods of this kind are only applicable to a limited subset of architectures and only when the trained model's internal state is accessible.

\subsection{Model-Agnostic Saliency Mapping Methods}

Fong et al.~\cite{Fong2017-jg} propose Extremal Perturbation, using gradient descent to learn a perturbation mask which minimises (or conversely, maximises) the prediction of the target class. This method produces binary segmentations, which look appealing, but obscure any difference in feature salience \emph{within the broadly salient region} of pre-specified size. Extremal Perturbation requires the selection of many hyperparameters (learning rate, number of iterations, mask upsampling factor, degree of blur, degree of jitter, et cetera) which are chosen empirically and may not generalise to novel models or datasets -- once more raising the lengthy and uncertain prospect of manual tuning to generate informative, interpretable saliency maps. The published settings for PASCAL VOC and COCO result in excellent performance, but take an extremely long time in comparison to other methods (over fifty seconds, compared to sub-second performance for most other methods). Other techniques consist of training a second model using a saliency criterion to generate attribution maps directly from the input sample~\cite{Ribeiro2016-xl, Dabkowski2017-td}, an approach which is very fast once the saliency model has been trained. However, applying this approach to a new dataset and model would require retraining the saliency model too, which may not be trivial -- necessitating not only training a predictive model to succeed at the task at hand, but also training a separate second model for saliency -- effectively doubling the hyperparameter and architecture tuning burden, and increasing computational cost. More importantly, explanations generated in this way are by their nature fundamentally divorced from the model in question, and invite biased tuning to generate saliency maps that \emph{look} sensible to humans, rather than optimising for explanatory power.

Other model-agnostic methods work by iteratively perturbing regions of the input sample~\cite{Petsiuk2018rise, Zeiler2013-lm}, and using the sensitivity of the model output to these perturbations to generate a saliency map. These methods have the nice property of direct interpretability (i.e.\ a perturbation in the input can be directly mapped to a change in the output), but are computationally expensive. Typically they also require heuristic parameter selection (for example, selecting the size of the perturbation kernel or number of masks to generate) to produce informative visualisations, which may necessitate many trials, and thereby also prove costly. Zeiler et al.~\cite{Zeiler2013-lm} outline an early form of this approach, in which a perturbation kernel of fixed size is iteratively applied to the input, and the difference in output at each kernel location is collated to form a saliency map. This is intuitive, but very time consuming, as it relies on running potentially many trials with different kernel dimensions to generate informative visualisations, since it is impossible to know the scale of the most salient features ahead of time. RISE~\cite{Petsiuk2018rise} is based on the same perturbation technique, and works by generating a number of low resolution random binary masks, upsampling them using bilinear interpolation, using them to mask the input, and weighting each mask by the model's output for the correspondingly masked input (the perturbation in this case being the dimming of the input). The weighted masks are then aggregated and normalised, producing a saliency map. The dimensions of the low resolution and upsampled masks, and the number of masks used, are chosen empirically (8000 masks were used for ResNet50). The fact that the masks are randomly generated means that RISE must always use a relatively large enough number of masks in order to avoid biasing the saliency map with unevenly distributed perturbations, especially when there are several salient regions of varying sizes contained in the input. The larger the input dimension, the larger the number of masks must be -- and crucially, one cannot really know ahead of time how many masks is sufficient to generate a faithful map. Decreasing the resolution of the initial binary mask before interpolation decreases the number of masks necessary -- however, the lower resolution this mask is, the coarser the final saliency map will be, making RISE prohibitively expensive for high resolution data which demands high resolution saliency. These limitations are explicitly mentioned in the original publication~\cite{Petsiuk2018rise}, which calls for future work to address this by intelligently selecting a smaller number of masks -- as we do here.

To address the limitations outlined above we propose a novel approach to perturbation-based saliency mapping which identifies salient regions regardless of scale, largely removes the need for heuristic parameter selection, and dramatically reduces computational cost whilst maintaining accurate saliency identification. We call this approach Hierarchical Perturbation (HiPe), as it provides these benefits by iteratively identifying salient sub-regions and disregarding relatively unimportant ones in increasing resolution. We compare our approach to other saliency mapping methods on the well established pointing game benchmark and the causal insertion/deletion metric~\cite{Petsiuk2018rise}.

\section{Proposed Method}
\begin{figure}
    \centering
    \includegraphics[width=1\textwidth]{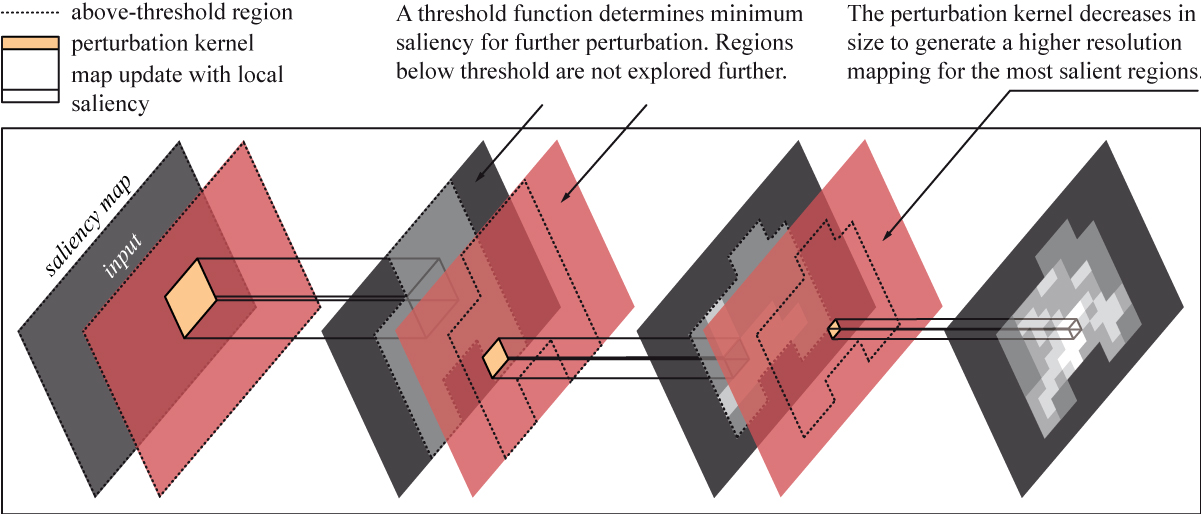}
    \caption{Saliency Mapping with Hierarchical Perturbation}
    \label{fig:hipe}
\end{figure}

 Hierarchical Perturbation is a natural extension of iterative occlusion~\cite{Zeiler2013-lm} and the random masking of RISE~\cite{Petsiuk2018rise} described in the previous section, in which we adopt the same principles of empirical salience estimation, but apply them in a more directed fashion to minimise computational cost and thereby make model-agnostic saliency mapping a realistic prospect for large samples or datasets. Our key insight is that a large amount of superfluous computation is performed when regions that have little effect on the model output are iteratively perturbed, or when random perturbation region selection results in spatially similar or overlapping regions. By avoiding this unnecessary cost through salience thresholding we are able to perform model-agnostic saliency mapping an order of magnitude faster than existing methods. As shown in Figure~\ref{fig:hipe}, HiPe does this by focusing on perturbing the most salient regions with increasing resolution whilst ignoring regions which do not not change the model’s output.

Let $f:x \rightarrow \mathbb{R}$ be our trained model, which takes $x$, a matrix of size $3 \times h \times w$ (in our case a three channel colour image with variable height and width), and returns a scalar confidence which we wish to attribute to a greater or lesser degree to some elements of $x$. We instantiate $s$ as our saliency map, initially a zero matrix of size $h \times w$, and populate it as follows:

We let our mask placeholder $m:\Lambda \rightarrow \{0,1\}$ be a zero matrix of size $d \times d$, where $d = \lceil \log_{2}(\min{(h,w)}) \rceil$. Using the rounded log of base two results in a mask that is neatly divisible by two, ensuring that the perturbation regions will evenly overlap. We use the minimum input dimension for convenience, as most real-life and benchmark data images do not stray too far from the square. For images with greater eccentricity it would be necessary to adjust this.

We then use a step function $t(s,m) \rightarrow \{0,1\}$ using the mid-range of the current saliency map as a threshold to identify regions of high salience for higher-resolution mapping thus (where $\circ$ denotes the Hadamard product):

\begin{equation}
    t(s,m) = \left\{\begin{matrix}
    1, & if & \max{(s\circ m)}\geq \min{s} + \frac{(\max{s} - \min{s})}{2}\\
    0, & otherwise &
    \end{matrix}\right.
\end{equation}

\begin{figure}
    \centering
    \includegraphics[width=0.8\textwidth]{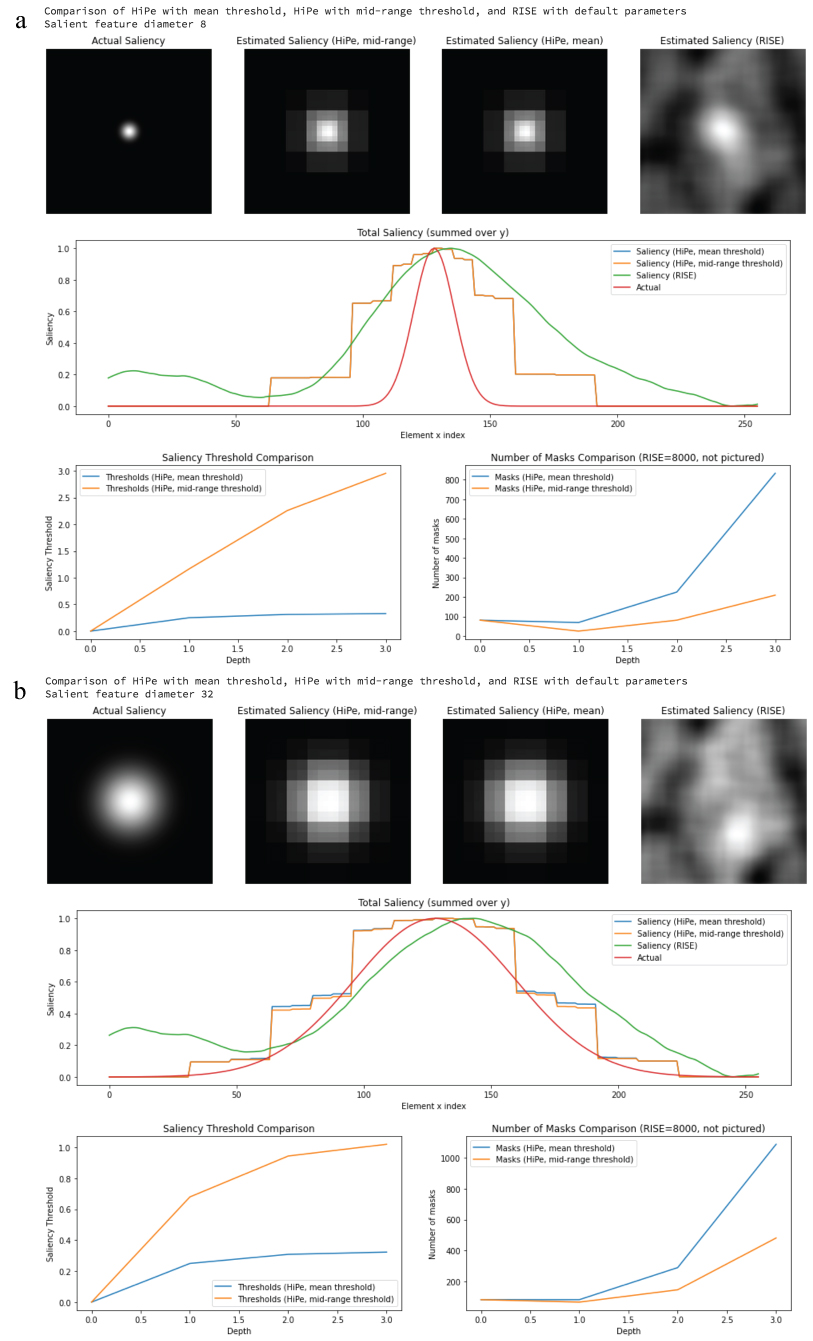}
    \caption{Here we generate some arbitrary $256 \times 256$ input data with a central blurred `salient' region of two sizes (diameter 4 above, and diameter 32 below), use a summation operation as a proxy model, and compare RISE with default parameters (and the default 8000 masks) against HiPe with two different saliency threshold methods. We show that for HiPe, using the mid-range of the current saliency map as the threshold generates comparable maps to using the mean, but does so far more efficiently, using higher thresholds and so requiring fewer masks. We also show that HiPe generates superior saliency maps with far fewer masks than RISE at both feature sizes.}
    \label{fig:hipe_comp}
\end{figure}

The mid-range of the current saliency map is used in preference to some arbitrary threshold as it allows us to handle varying saliency distributions across different samples. Selecting an optimal threshold for all samples would require either knowledge about the saliency distribution beforehand, or costly heuristic trials -- whilst manually selecting one for each sample would optimise for achieving visually pleasing saliency maps, and render comparing the relative saliency of different samples across the dataset difficult. We use the mid-range for this rather than the mean or some other aggregate of the current map, as the mid-range is extremely sensitive to outliers. This sensitivity allows us to focus on the most salient regions extremely quickly in cases where only a small region of the input is salient, greatly improving performance compared to the mean. An example of this is shown in Figure~\ref{fig:hipe_comp}.

Note that the first time this thresholding operation is applied, the saliency map consists only of zeros, and so $t(s,m) = 1$ in every case. We define the set of perturbation masks $M$ such that

\begin{equation}\{m \in M | \sum{m_{i:i+2,j:j+2}} = 4 \wedge t(s,m) = 1\}\end{equation} so all masks which contain a $2 \times 2$ region of ones, with all other elements set to zero (this is to ensure an overlap of perturbation to capture features on the border of regions), and meet the threshold. For each mask we \emph{perturb} a region of the input (which we might think of as the region of $x$ which corresponds to the non-zero region of each $m \rightarrow M$), by replacing all pixels therein with the mean of that region, such that (with scaling factors $\sigma = \lfloor\frac{H}{d}\rfloor, \omega = \lfloor\frac{W}{d}\rfloor$): 

\begin{equation}
    x' = x|x_{i\cdot\sigma:(i+2)\cdot\sigma,j\cdot\omega:(j+2)\cdot\omega} = \sum{\frac{x_{i\cdot\sigma:(i+2)\cdot\sigma,j\cdot\omega:(j+2)\cdot\omega}}{4\sigma\omega}}
\end{equation}

This usage of the local mean as a perturbation substrate is further discussed in section 5. We upsample $M$ to size $h \times w$ using proximal interpolation -- note that we do not artificially smooth the mask during upsampling -- and update our saliency map $s$ such that: 

\begin{equation}
    s = s + ReLU(f(x)-f(x')) \circ |M-1|
\end{equation}

We use ReLU so that we are thresholding only with respect to perturbations which decrease the confidence of the target class, and therefore only highlight regions which, when available to the model, have a positive influence on the target class confidence. We then double $d$, and repeat the above while $d \leq \frac{\min{(h,w)}}{4} \wedge \#M > 0$. This effectively means that HiPe can capture features as small as $2\times2$ pixels in size, as the minimum perturbation region size is $4 \times 4$, with a 50\% overlap.

To summarise intuitively, HiPe begins by perturbing large, overlapping regions and using the difference in the model output for each of these perturbations to update the saliency map. All regions of the saliency map which exceed the saliency threshold are split into smaller overlapping regions, which are then each perturbed, and the saliency map updated in turn -- and so on -- until either the minimum perturbation size is reached, or no region remains above the saliency threshold. By discarding regions with little impact on the model's output and focusing only on the more salient areas, we are able to generate saliency maps of comparable quality to the state-of-the-art for model-agnostic methods, at a fraction of the computational cost.  

\section{Experiments}

In this section we evaluate our method and compare it with popular alternatives from the literature using two widely used saliency mapping benchmarks -- the pointing game and insertion/deletion causal metrics. As raised by Zeiler et al.~\cite{Zeiler2013-lm} we believe the pointing game is something of a flawed metric -- it relies on the assumption that if a model is good, it is good because it learns the same features that a human would, and so a good saliency map would highlight those same features. This means that it rewards maps that encode what a human being would consider to be salient, rather than what is in fact salient to the model which we hope to understand. For example, a model classifying a fridge may also identify a microwave as equally salient, as it provides important contextual information which may contribute a large part of the `fridge' confidence. A saliency method faithfully identifying this would be penalised for it. We \emph{cannot} assume that even an 100\% accurate model has learned to use the same features that we would to make a prediction, and so must beware of biasing the development of saliency methods to confirm our assumptions, rather than provide accurate characterisations of what the model is really doing. 

Nonetheless, it appears that the pointing game is at least a reasonable proxy, given a well trained model, and since it is a standard comparator we include it here, along with the more objective insertion and deletion metrics~\cite{Petsiuk2018rise}. We use a publicly available \href{https://pytorch.org/docs/stable/torchvision/models.html}{pre-trained ResNet50 model}~\cite{He}, the \href{https://cocodataset.org/}{MSCOCO 2014} validation set and the \href{http://host.robots.ox.ac.uk/pascal/VOC/voc2007/}{VOC 2007} test set, to allow for comparison with previous works.

\subsection{Pointing Game}

The pointing game measures the accuracy of a given saliency map by examining the correlation between the most salient point on that map with the location of the object in question. This is done by generating a saliency map for some class prediction for a given image, and comparing the location of the semantic annotation of the class object with the maximum (therefore most salient point) on that map. If the maximum point falls within the boundary of the object annotation, one point is gained, and the overall accuracy is calculated as the number of hits divided by the number of hits plus misses.

\begin{table}
    \centering
    \begin{tabular}{| c c | c c | c | c c | c |}
        \hline
        \multicolumn{2}{| c |}{}
         & \multicolumn{3}{c |}{COCO14 Val}
         & \multicolumn{3}{c |}{VOC07 Test} \\
        \hline
        & Method & All & Diff & Time (s) & All & Diff & Time (s) \\
        \hline
        & cMWP~\cite{Zhang2016-vm} & 58.5 & 53.6 & 0.08 & 90.6 & 82.2 & 0.09 \\
        & GCAM~\cite{Selvaraju2016-vq} & 57.3 & 52.3 & 0.03 & 90.4 & 82.3 & 0.03 \\
        \hline
        \multirow{3}{*}{MA}
        & ExtP~\cite{Fong2019-wy} & \textbf{55.7} & 46.9 & 53.4 & 86.3 & 73.4 & 53.5 \\
        & RISE*~\cite{Petsiuk2018rise} & 55.6 & -- & 25.87 & \textbf{88.9} & -- & 23.61 \\
        & \textbf{HiPe} & 54.6 & \textbf{49.6} & \textbf{0.94} & 85.6 & \textbf{75.1} & \textbf{0.95} \\
        \hline
        & MWP~\cite{Zhang2016-vm} & 49.6 & 43.9 & 0.06 & 84.4 & 70.8 & 0.06 \\
        & Guid.~\cite{Springenberg2014-rv} & 42.1 & 35.3 & 0.1 & 77.2 & 59.5 & 0.1 \\
        & Grad~\cite{Simonyan14a} & 35.0 & 29.4 & 0.06 & 72.3 & 56.8 & 0.06 \\
        & DConv~\cite{Zeiler2013-lm} & 30.0 & 21.9 & 0.06 & 68.6 & 44.7 & 0.06 \\
        \hline
    \end{tabular}
    \caption{\textbf{Pointing Game}: Mean accuracy and time per sample on full and difficult validation/test sets of VOC07 and COCO14 (as defined by Zhang et al.~\cite{Zhang2018-ji}), using contemporary model-specific and model-agnostic (MA) methods. We highlight superior results within the model-agnostic subset. Accuracy results for methods other than our own are taken from Fong et al.~\cite{Fong2019-vk}, whereas average saliency map generation times are obtained using \href{https://github.com/facebookresearch/TorchRay}{Torchray} implementations with default settings for 1000 random samples (one random class per sample) using an NVIDIA GeForce RTX 2060. *RISE results taken  from Petsuik et al.~\cite{Petsiuk2018rise} which excluded the difficult subset.}
    \label{tab:pointing_game}
\end{table}

Table~\ref{tab:pointing_game} shows that HiPe is competitive with existing model-agnostic saliency mapping methods Extremal Perturbation (ExtP) and RISE, whilst taking on average, under a second to produce a saliency map. In contrast, RISE requires over 23 seconds, and Extremal Perturbation nearly a minute per image. Both of these methods could be made faster by decreasing the number of random masks, or the number of iterations respectively -- but at the cost of increasing noise in the saliency map generated. 

\subsection{Insertion and Deletion Metrics}

\begin{figure}
    \centering
    \includegraphics[width=1\textwidth]{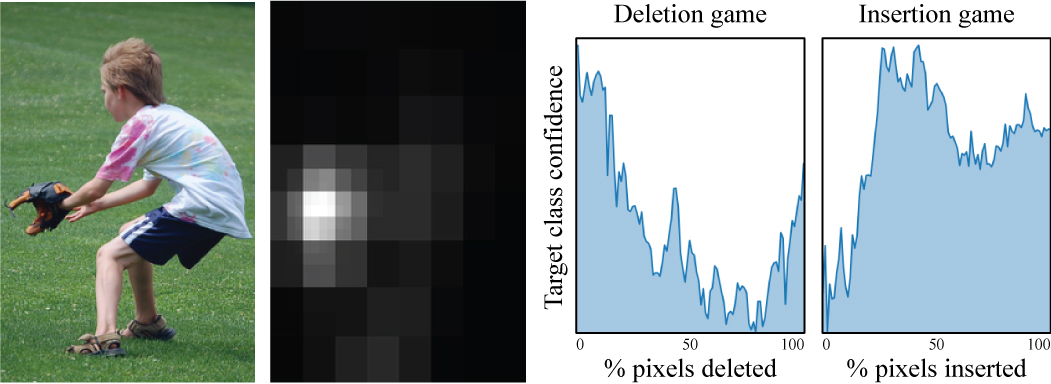}
    \caption{Example of the insertion and deletion metrics using HiPe, for the class `baseball glove'. The area under the curve (AUC) is used to benchmark the accuracy of the saliency maps -- lower is better for deletion, higher is better for insertion.}
    \label{fig:ins_del}
\end{figure}

Proposed by Petsuik et al.~\cite{Petsiuk2018rise}, the idea behind the insertion and deletion metrics is that, intuitively, the more salient pixels we show the model, the more confident its prediction should be -- and conversely, the more salient pixels we remove from the input, the more the confidence should drop. If the pixels identified as salient by some method are indeed the same pixels that most influence a model's output, we would expect to see a sharp drop in confidence and a low AUC as they are removed in decreasing order of salience. Likewise, if we begin with an entirely obscured input and \emph{introduce} pixels in decreasing order of salience, we would expect a sharp increase in confidence and a large AUC -- see Figure~\ref{fig:ins_del}. Unlike the pointing game, these metrics are self-contained and unbiased by human assumptions, and approximate only whether the ascribed saliency values per pixel map accurately to the change in model output. For our experiments, the percentage of pixels removed or added at each step is set to 1\%, we use the blurred substrate for insertion, and the zero substrate for deletion as per the literature.

\begin{table}
    \centering
    \begin{tabular}{| c c | c c | c c |}
    \hline
    \multicolumn{2}{| c |}{}
     & \multicolumn{2}{c |}{COCO14 Val}
     & \multicolumn{2}{c |}{VOC07 Test} \\
    \hline
    & Method & Insertion & Deletion & Insertion & Deletion \\
    \hline
    MA & \textbf{HiPe} & \textbf{0.68} & 0.43 & \textbf{0.67} & 0.42 \\
    \hline
    & GCAM & 0.67 & 0.41 & 0.67 & 0.39 \\
    & Guid. & 0.66 & 0.39 & 0.65 & 0.38 \\
    & cMWP & 0.66 & 0.39 & 0.65 & 0.38 \\
    \hline
    \multirow{2}{*}{MA}
    & RISE* & 0.65 & \textbf{0.40} & 0.65 & \textbf{0.39} \\
    & ExtP & 0.63 & 0.45 & 0.62 & 0.45 \\
    \hline
    & MWP & 0.62 & 0.38 & 0.62 & 0.34 \\
    & DConv & 0.62 & 0.39 & 0.62 & 0.44 \\
    & Grad & 0.62 & 0.46 & 0.61 & 0.44 \\
    \hline
    \end{tabular}
    \caption{\textbf{Insertion and Deletion AUC:} mean Area Under Curve for insertion (higher is better) and deletion (lower is better) causal metrics described by Petsuik et al.~\cite{Petsiuk2018rise} on 1000 randomly selected samples. As in Table~\ref{tab:pointing_game} we highlight superiority within the subset of model-agnostic (MA) methods.}
    \label{tab:causal_metrics}
\end{table}

In Table~\ref{tab:causal_metrics} we see that HiPe is competitive with all methods across both metrics and datasets, and \textit{outperforms} all methods on the insertion metric. We would expect this to be the case, because unlike the model-specific methods, the saliency maps generated by HiPe are products of directly perturbing the input, in a comparable way to the causal metrics we use here. This is also the case for RISE -- the small decrease in comparative performance is due to the stochasticity inherent in this method.

\begin{figure}
    \centering
    \includegraphics[width=1\textwidth]{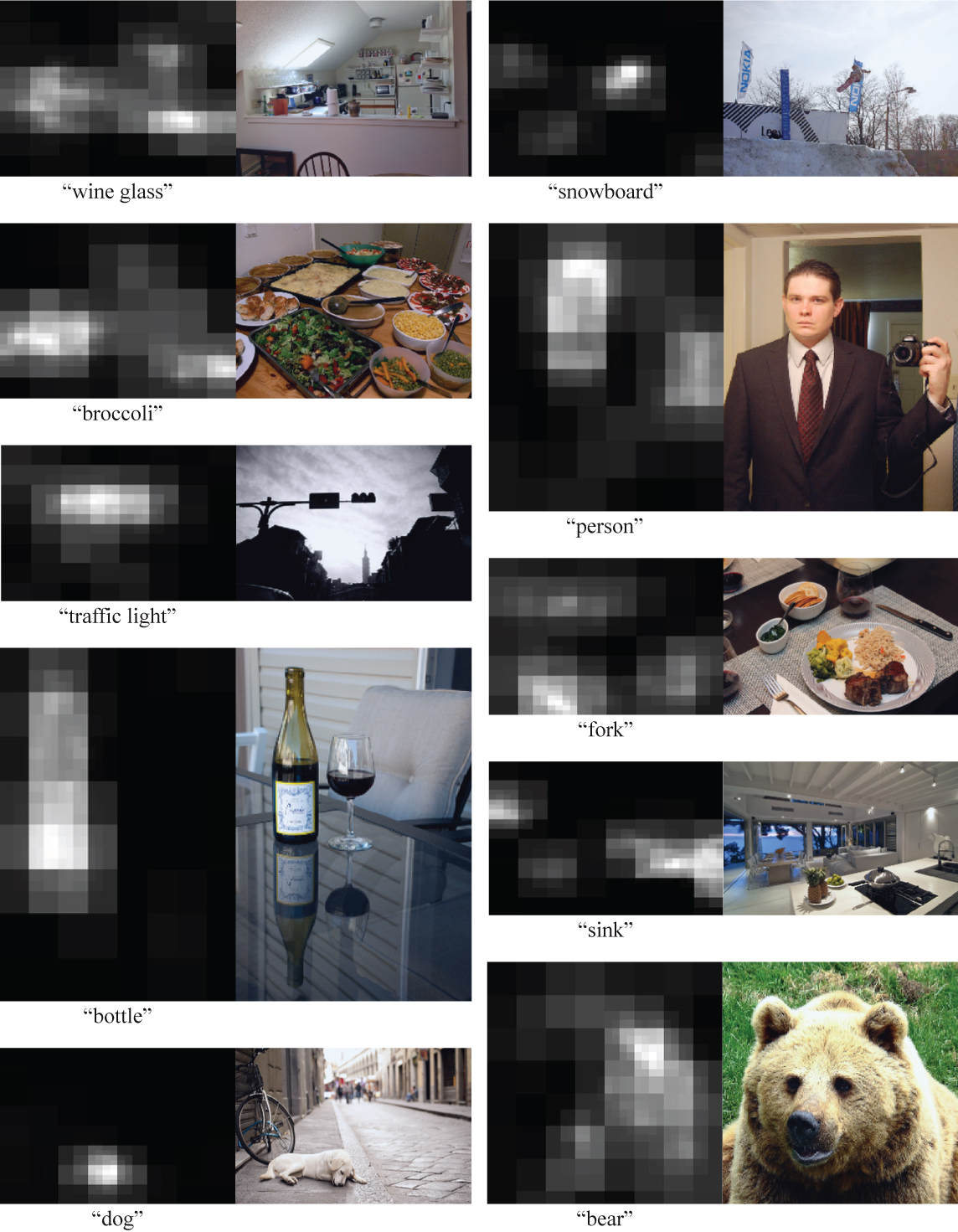}
    \caption{Examples of saliency maps generated with Hierarchical Perturbation. Note that more salient regions are of higher resolution.}
    \label{fig:hipe_examples}
\end{figure}

\begin{figure}
    \centering
    \includegraphics[width=1\textwidth]{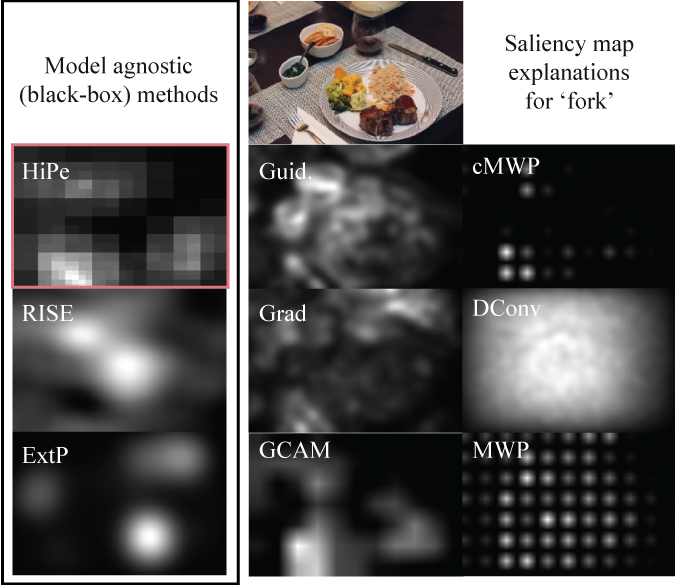}
    \caption{Examples of saliency maps for the class `fork'. This example was chosen to highlight the surprising variation in maps generated with different methods.}
    \label{fig:hipe_fork}
\end{figure}

Figure~\ref{fig:hipe_examples} shows examples of saliency maps generated with HiPe, and Figure~\ref{fig:hipe_fork} compares saliency maps across all methods for a single image, for the class `fork'. We can see that HiPe, Grad-CAM and cMWP identify the fork in the image as salient, but the other methods do not. It has been shown that deconvnet and guided backpropagation are in some cases invariant to reparameterization in later layers~\cite{Nie2018-zw}, and essentially act as image reconstructors -- this is particularly evident in the guided backpropagation (Guid.) map here. We suspect that the failure of RISE and Extremal Perturbation to localise the fork in the image is due to the small size of the features in question -- both RISE and Extremal Perturbation rely on heuristic hyperparameters which dictate the size of salient regions to be localised. HiPe, by contrast, performs well on inputs containing salient regions of all sizes.

\begin{figure}
    \centering
    \includegraphics[width=1\textwidth]{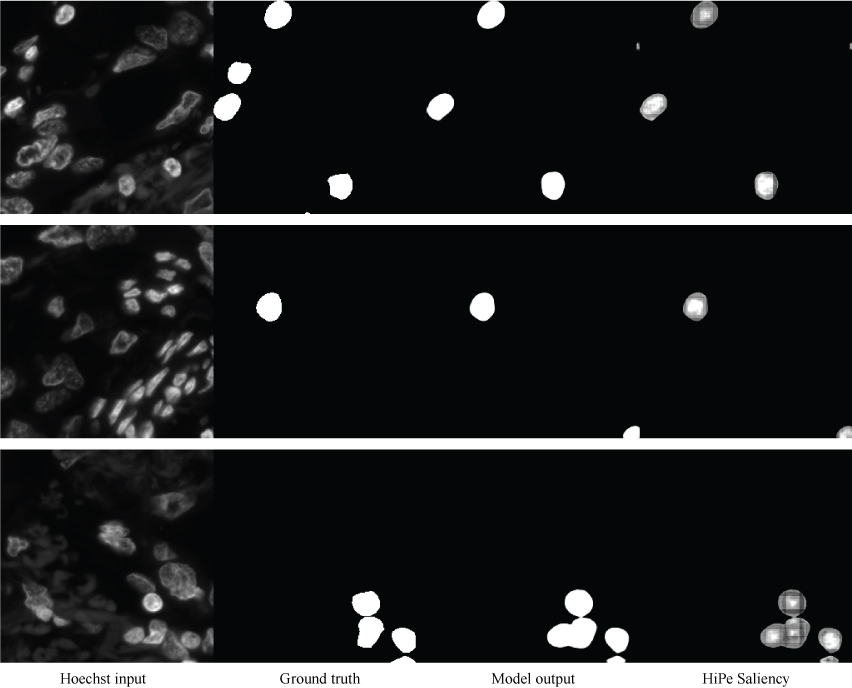}
    \caption{Use of HiPe for an immune cell segmentation task using a deep residual U-net on Hoechst-stained biopsy slides. Here HiPe enables fast and detailed saliency map generation for high resolution images with small, sparse features.}
    \label{fig:seg_example}
\end{figure}

\begin{figure}
    \centering
    \includegraphics[width=1\textwidth]{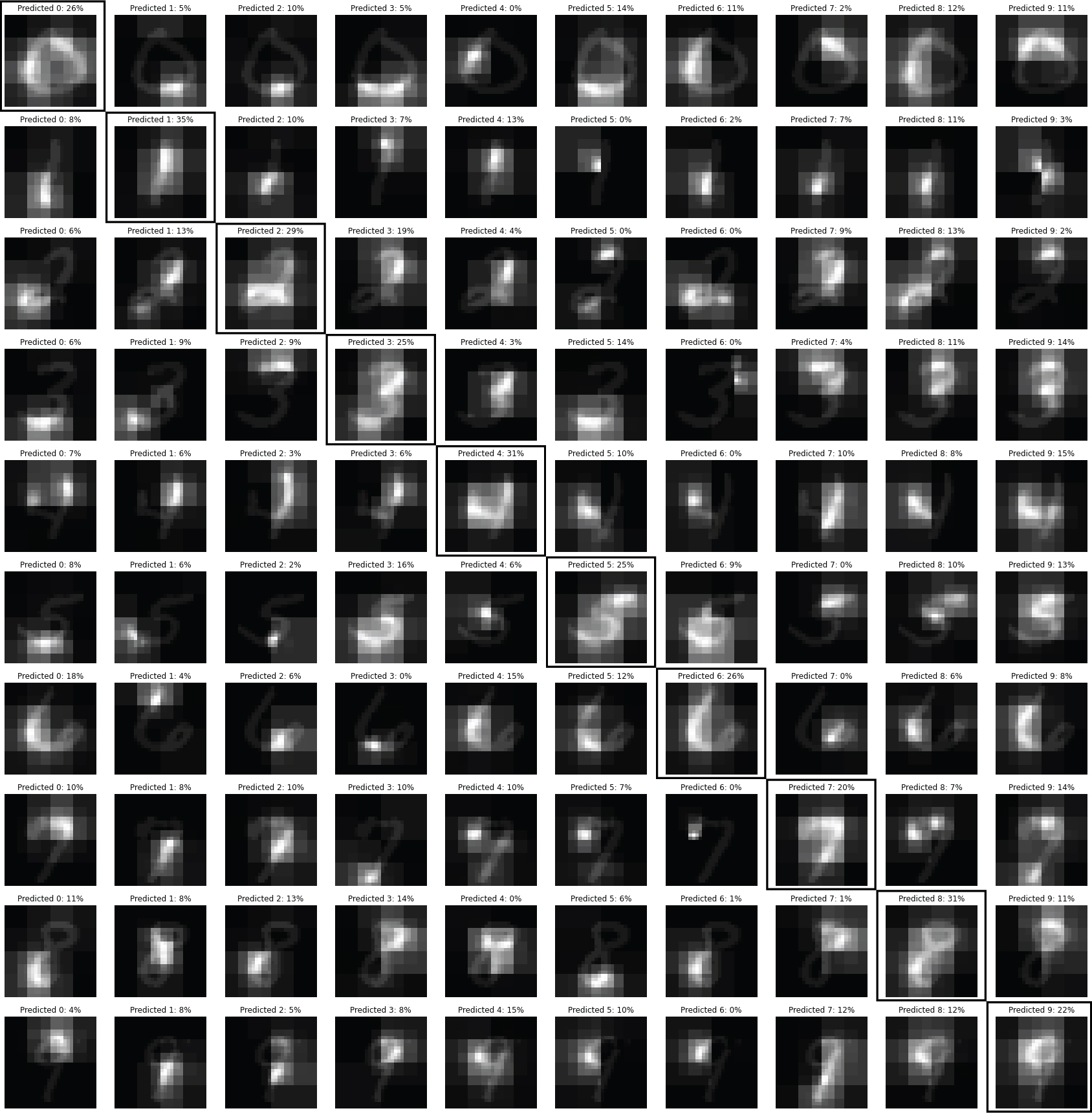}
    \caption{HiPe saliency on MNIST for each class. The left-to-right diagonal contains the saliency of the correct class for each sample. The probabilities shown here are the model output logits normalised to sum to one.}
    \label{fig:hipe_vis_mnist}
\end{figure}

Figure~\ref{fig:seg_example} shows an example of HiPe applied to a segmentation task -- that of immune cell segmentation from Hoechst-stained Whole Slide Images (WSIs) using a deep residual U-net~\cite{Cooper2021-xq}. This example showcases the applicability of our model-agnostic method to arbitrary architectures, and leverages the speed and robustness of HiPe in a use-case where other perturbation-based techniques would have been prohibitively slow for multiple large WSI images, and gradient-based methods too indistinct for the relatively small and sparse salient features (as shown in our benchmark experiments -- see Figure~\ref{fig:hipe_fork}).

As raised by Adebayo et al.~\cite{Adebayo2018-mj} and Kindermans et al.~\cite{Kindermans2019-vv}, robust saliency mapping algorithms must be sensitive to input in respect to their target output class -- for example, given an image containing both a cat and a dog, and a trained neural network that classifies cats and dogs we would expect that a saliency map generated with respect to the 'cat' output would look very different to that of 'dog' -- given of course, that the neural network has successfully learned 'cat' and 'dog' specific features.

In order to investigate this property, we generate \textit{class-specific} HiPe saliency maps on the MNIST dataset. We choose MNIST in this case partly because it a common sanity-checking dataset in the literature and therefore allows for easy comparison with other saliency-mapping experiments~\cite{Kindermans2019-vv}, but primarily because due to its simplicity it is free from potentially confounding spurious correlations in the input. Therefore we can confidently expect a well-trained model to classify digits based only on the digits themselves.

We train a simple three layer convolutional network using cross entropy loss and SGD with a learning rate of 0.001 and momentum of 0.9. The network had two convolutional layers (the first having 16 channels, and the second 32), each with a kernel size of 3, a stride of 1, and zero padding. Each was followed with a ReLu and maximum pooling with kernel size 2. The final was a standard linear layer. We train this network to 97\% accuracy, and apply HiPe to each class output in turn. Figure~\ref{fig:hipe_vis_mnist} shows the results of this experiment on a randomly chosen sample for each digit. The saliency maps for each digit are distinct and sensible. This image confirms that HiPe is sensitive to target class for each map, and can be interpreted intuitively -- we see that the zero input has high saliency for the zero class around the entirety of the digit. For other class saliency maps on the zero input, HiPe finds portions of the input salient -- for example, the upper curve for class 9, the lower curve for classes 3 and 5, and the leftmost curve for class 6.

\section{Discussion}

Unlike RISE~\cite{Petsiuk2018rise} which generates masks randomly, and methods which learn secondary models through gradient descent such as those of Fong et al., Ribeiro et al. and Dabwoksi et al.~\cite{Fong2019-vk,Ribeiro2016-xl, Dabkowski2017-td}, HiPe contains no random elements. However, HiPe is not able to capture instances where the salience of two spatially distinct features in combination is greater than the sum of each feature individually (i.e.\ if perturbing the microwave and oven \emph{at the same time} changes the prediction more than the difference when perturbing the oven \emph{plus} the difference when perturbing the microwave), since we perturb only one locality at a time -- this unlike RISE, which with a large enough number of random masks will capture this phenomenon, and unlike Extremal Perturbation, which by design will capture any combination of features if trained for long enough (albeit at significant time cost, as our experiments show). 

\begin{figure}
    \centering
    \includegraphics[width=1\textwidth]{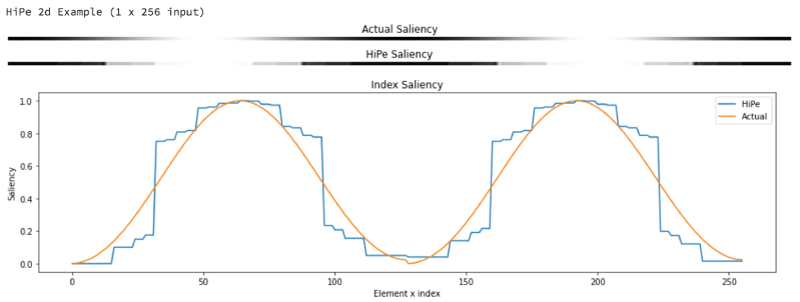}
    \caption{A toy example of HiPe applied to non-image data. In this case, we generated a 2d array of values to represent the actual saliency, using summation as a proxy `model', and used HiPe to estimate the true salience.}
    \label{fig:hipe_2d}
\end{figure}

As we have discussed, HiPe is model-agnostic and as such can be applied to any model which maps an input to an output. This includes data of arbitrary dimension -- with minor adaptation HiPe could be applied to data of any dimension in which the spatial relationship between input elements has predictive power. This includes two-dimensional time-series data, four dimensional video, or graph data, et cetera. Figure~\ref{fig:hipe_2d} shows a toy example of this, using HiPe to identify the most salient features in two-dimensional input data using a proxy model which simply sums the input, as in Figure~\ref{fig:hipe_comp}. 

The choice of perturbation substrate is likely important in perturbation based saliency mapping~\cite{Brunke2021-cr}, although we find empirically that using our method, the local mean of the perturbed region, results in marginally superior performance (around 1\% on the pointing game benchmark) to a zero substrate (equivalent to the dataset mean, since inputs are standardised during pre-processing) or to a Gaussian blurred substrate. Blurred, noisy or zero substrates are the most commonly used~\cite{Fong2019-vk} -- we might add local mean to that list, given the results herein. It is possible to use any kind of perturbation technique or substrate with HiPe -- this is useful, because \emph{how} best to perturb the input for perturbation-based saliency mapping methods remains an open question~\cite{Brunke2021-cr}.

HiPe does not apply any smoothing, either to the perturbed region (which is typically done to make the perturbation appear more natural), or to the resulting saliency map (which some methods do purely for the sake of visual appeal rather than for a fundamental theoretical reason). We found that using a similar method to RISE in which a low resolution mask is upsampled using bilinear interpolation in order to generate smooth perturbations actually resulted in a small decrease in performance when using HiPe -- this is in contrast to RISE and Extremal Perturbation, which explicitly apply smoothing. A possible hypothesis for the appeal of smooth perturbations, and resulting smooth saliency maps, is simply human visual preference (we assume sharp artefacts are more visually disturbing than smooth ones~\cite{Fong2019-vk}) rather than explanatory power -- we may assume that smooth perturbations are less confounding, but this is not borne out by our results.

\section{Conclusion}

Most state of the art work in saliency mapping is limited by model-specificity, assuming a certain subset of architectures and requiring on access to the internal state of the model. The few model-agnostic methods that do exist perform well on the benchmarks we have examined, but can be very slow -- prohibitively so for large samples. To address this, we have presented Hierarchical Perturbation (HiPe), a fast and easily interpretable explanatory algorithm for understanding arbitrary models, regardless of their architecture. HiPe is able to create model-agnostic saliency maps so quickly because it is content-aware in a way that existing perturbation-based saliency mapping algorithms are not. Petsuik et al., Fong et al., and Zeiler et al.\cite{Petsiuk2018rise, Fong2017-jg, Zeiler2013-lm} require a pre-specified number of iterations -- whether that is epochs, number of random masks generated, or occlusion kernel size and step -- which fix the amount of computation required for an input of given size irrespective of the proportion of the input that is actually salient. It is also impossible to know ahead of time what the optimal value for these parameters might be in order to trade-off accuracy and efficiency, and finding the optimal parameters (for one input sample, or across an entire dataset) may require many trials. Additionally, these parameters limit the size of salient region that can be detected, which can lead to omissions as in Figure~\ref{fig:hipe_fork}. Our method, by contrast, continually disregards regions which have little impact on the model output, and by so doing so inherently limits the amount of computation required without imposing restrictions on the size of the salient region it is possible to detect. We also note that HiPe is data-agnostic, as well as model-agnostic -- although we benchmark it on images here to allow for comparison with existing methods, it may be applied to data of arbitrary dimension. We have shown that the maps generated by HiPe are of comparable quality to state-of-the-art saliency mapping methods, but require a fraction of the computational cost compared to other model-agnostic approaches, and as such we expect HiPe to prove invaluable to researchers and practitioners working with models of all kinds.

A python implementation of HiPe and our benchmarking scripts are available \href{github.com/jessicamarycooper/Hierarchical-Perturbation}{here}.

\section{Acknowledgements}
This work is supported by the Industrial Centre for AI Research in Digital Diagnostics (iCAIRD) which is funded by Innovate UK on behalf of UK Research and Innovation (UKRI) [project number: 104690].

\bibliography{refs}
\end{document}